\documentclass[10pt,journal,compsoc]{IEEEtran}
\ifCLASSOPTIONcompsoc
  \usepackage[nocompress]{cite}
\else
  \usepackage{cite}
\fi

\usepackage{graphicx}

\ifCLASSINFOpdf
\else
\fi
\usepackage[utf8]{inputenc} 
\usepackage[T1]{fontenc}    
\usepackage{url}            
\usepackage{booktabs}       
\usepackage{amsfonts}       
\usepackage{nicefrac}       
\usepackage{microtype}      
\usepackage{graphicx}
\usepackage{amsthm}
\usepackage{amsmath}
\usepackage{wrapfig}
\usepackage{xcolor}
\usepackage{tabularx}
\usepackage[mathscr]{euscript}
\usepackage{multicol}

\newcolumntype{Y}{>{\centering\arraybackslash}X}

\usepackage{multirow}
\usepackage{hhline}
\usepackage{array}
\usepackage{amsmath}
\usepackage{amssymb}
\usepackage{bm}
\usepackage{color}
\usepackage[colorlinks = true,
            linkcolor = blue,
            urlcolor  = blue,
            anchorcolor = blue]{hyperref}
\definecolor{COLOR}{rgb}{1.0, 0.25, 0.25}

\begin{document}
%
\title{Topology-Aware Graph Pooling Networks}
%
%
%
%

\author{Hongyang~Gao*,
        Yi~Liu*,
        and~Shuiwang~Ji,~\IEEEmembership{Senior Member,~IEEE}
\thanks{Hongyang Gao is with the Department
of Computer Science, Iowa State University, Ames, IA 50011, USA
(email: hygao@iastate.edu).}
\thanks{Yi Liu and Shuiwang Ji are with the Department
of Computer Science \& Engineering, Texas A\&M University, College Station, TX 77843, USA
(email: yiliu@tamu.edu, sji@tamu.edu).}
\thanks{*These authors contributed equally to this work.}}

\IEEEtitleabstractindextext{%
\begin{abstract}
Pooling operations have shown to be effective on computer vision and natural language processing tasks. One challenge of performing pooling operations on graph data is the lack of locality that is not well-defined on graphs. Previous studies used global ranking methods to sample some of the important nodes, but most of them are not able to incorporate graph topology. In this work, we propose the topology-aware pooling (TAP) layer that explicitly considers graph topology. Our TAP layer is a two-stage voting process that selects more important nodes in a graph. It first performs local voting to generate scores for each node by attending each node to its neighboring nodes. The scores are generated locally such that topology information is explicitly considered. In addition, graph topology is incorporated in global voting to compute the importance score of each node globally in the entire graph. Altogether, the final ranking score for each node is computed by combining its local and global voting scores. To encourage better graph connectivity in the sampled graph, we propose to add a graph connectivity term to the computation of ranking scores. Results on graph classification tasks demonstrate that our methods achieve consistently better performance than previous methods.
\end{abstract}

}

\maketitle

\IEEEdisplaynontitleabstractindextext

%
\IEEEpeerreviewmaketitle

\IEEEraisesectionheading{\section{Introduction}\label{sec:introduction}}

\IEEEPARstart{P}{ooling} operations have been widely applied in various
fields~\cite{simonyan2014very,he2016deep,huang2017densely,zhang2015character}.
Pooling operations can effectively reduce dimensional
sizes~\cite{simonyan2014very,mou2016convolutional} and enlarge
receptive fields~\cite{chen2017deeplab}. However, it is
challenging to perform pooling operations on graph data. In
particular, there is no spatial locality information or order
information among the
nodes~\cite{ying2018hierarchical,gao2018large,wang2020second}. Some works try
to overcome this limitation with two categories; those are node
clustering~\cite{ying2018hierarchical,yuan2019structpool} and node
sampling~\cite{gao2019graph,zhang2018end}. Node clustering
methods create graphs with super-nodes. The adjacency matrix of
the learned graphs in node clustering methods is softly
connected. These methods suffer from the over-fitting problem and
need auxiliary link prediction tasks to stabilize the
training~\cite{ying2018hierarchical}. The node sampling
methods like top-$k$ pooling~\cite{gao2019graph,zhang2018end}
rank the nodes in a graph and sample top-$k$ nodes to form the
sampled graph. It uses a small number of additional trainable
parameters and is shown to be more powerful~\cite{gao2019graph}.
However, the top-$k$ pooling layer does not explicitly
incorporate the topology information in a graph when computing
ranking scores, which may cause performance loss.

In this work, we propose a novel topology-aware pooling~(TAP)
layer that explicitly encodes the topology information when
computing ranking scores. 
Our TAP layer performs a two-stage voting process to 
examine both the local and global importance of each node. 
We first perform local voting and use dot product to compute
similarity scores between each node and its neighboring nodes.
The average similarity score of a node is used as its importance
score within a local neighborhood.
In addition, we perform global voting to weigh the importance of each
node globally in the entire graph.
The final ranking score for each node is
the combination of its local and global voting scores.
To avoid isolated nodes problem
in our TAP layer, we further propose a graph connectivity term
for computing the ranking scores of nodes. The graph connectivity
term uses degree information as a bias term to encourage the
layer to select highly connected nodes to form the sampled graph.
Based on the TAP layer, we develop topology-aware pooling
networks for network embedding learning. Experimental results on
graph classification tasks demonstrate that our proposed networks
with TAP layers consistently outperform previous models. The
comparison results between our TAP layer and other pooling layers
based on the same network architecture demonstrate the
effectiveness of our method compared to other pooling methods.


\section{Background and Related Work}\label{sec:bg}

The pooling operations on graph data mainly include two categories;
those are node clustering and node sampling.
DIFFPOOL~\cite{ying2018hierarchical} realizes graph pooling
operation by clustering nodes into super-nodes. By learning an
assignment matrix, DIFFPOOL softly assigns each node to different
clusters in the new graph with specified probabilities. The pooling
operations under this category retain and encode all node
information into the new graph. One challenge of methods in this
category is that they may increase the risk of over-fitting by
training another network to learn the assignment matrix. In
addition, the new graph is mostly connected where each edge value
represents the strength of connectivity between two nodes. The
connectivity pattern in the new graph may greatly differ from that
of the original graph.

The node sampling methods mainly select a
fixed number $k$ of the most important nodes to form a new graph. In
SortPool~\cite{zhang2018end}, the same feature of each node is used
for ranking and $k$ nodes with the largest values in this feature
are selected to form the coarsened graph. Top-$k$
pooling~\cite{gao2019graph} generates the ranking scores by using a
trainable projection vector that projects feature vectors of nodes
into scalar values. $k$ nodes with the largest scalar values are
selected to form the coarsened graph. These methods involve none or
a very small number of extra trainable parameters, thereby avoiding
the risk of over-fitting. However, these methods suffer from one
limitation that they do not explicitly consider the topology
information during pooling. Both SortPool and top-$k$ pooling 
rely on a global voting process but fail to
consider local topology information.
In this work, we propose a pooling operation
that explicitly encodes topology information in ranking scores,
thereby leading to an improved operation.

\section{Topology-Aware Pooling Layers and Networks}

\begin{figure*}[t]
    \centering
    \includegraphics[width=0.7\textwidth]{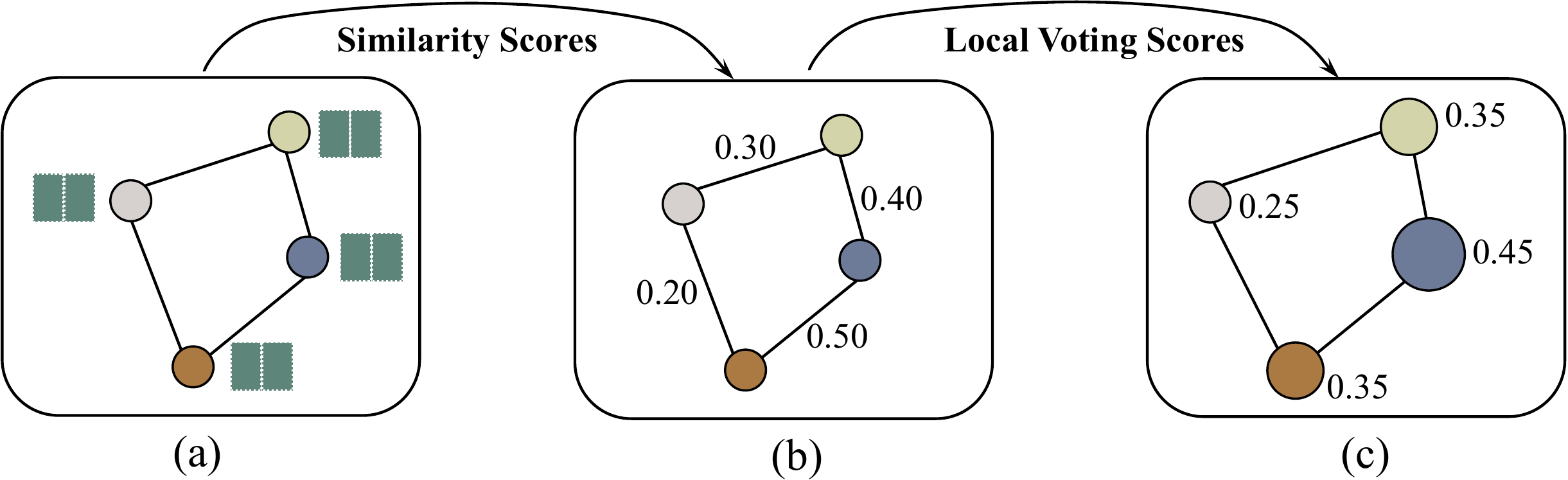}
    \caption{An illustration of the proposed local voting. This graph contains
    four nodes, each of which has 2 features. Given the input
    graph, we first compute similarity scores between every pair
    of connected nodes. In graph (b), we label each edge by the
    similarity score of its two end nodes. Then we compute the
    local voting score of each node by averaging the similarity scores
    with its neighboring nodes. In graph~(c), we label each node
    by its local voting score and a bigger node indicates a higher
    score.}\label{fig:gap}
\end{figure*}

In this work, we propose the topology-aware pooling~(TAP) layer
that encodes topology information in ranking scores for node
selection. We also propose a graph connectivity term in the
computation of ranking scores, which encourages better graph
connectivity in the coarsened graph. Based on our TAP layer, we
propose the topology-aware pooling networks for graph
representation learning.

\subsection{Topology-Aware Pooling Layer}\label{sec:gap}

\subsubsection{Graph Pooling via Node Sampling} \label{sec:graphpooling}
Pooling operations are important for deep models on image and NLP tasks that
they help enlarge receptive fields and reduce computational cost.
They are locality-based operations that extract high-level features
from local regions. When generalizing GNNs and GCNs to graph structured data,
graph pooling is an important yet challenging topic.
Recently, two families of graph pooling methods are proposed. One treats the graph pooling as a node clustering problem while the other one treats graph pooling as a node sampling problem. Both of them are developed to
scale down the size of node representations and learn new representations.
Formally,
given a graph
$G=(A,H)$ with the adjacency matrix $A\in \{0,1\}^{n\times n}$ and
the feature matrix $H\in\mathbb{R}^{n\times d}$, graph pooling produces a new graph $G^\prime$ with $k$ nodes. The new graph $G^\prime$ can be represented by its adjacency matrix $A^\prime\in \{0,1\}^{k\times k}$ and feature matrix $H^\prime\in\mathbb{R}^{k\times d}$. In this work, we follow~\cite{zhang2018end,gao2019graph} and treat the graph pooling as a node sampling problem, which learns the importance of different nodes in the original graph $G$ and samples the top $k$ important nodes to form the new graph $G^\prime$. Existing methods~\cite{zhang2018end,gao2019graph} generate node importance scores based on a
globally-shared projection vector that projects feature vectors to scalars.
Each scalar is an importance score indicating the importance of the corresponding node. 
However, these methods do not
explicitly consider graph topology information when performing graph pooling,
thereby leading
to constrained network capability.

In this section, we propose the topology-aware pooling~(TAP)
layer that performs nodes sampling by considering the
graph topology. 
Specifically, the node ranking score is obtained from two voting processes: local voting
and global voting.
For any node, the local voting computes its importance based
on the similarities with its neighboring nodes,
while the global voting evaluates its global importance in the entire graph.
Altogether, these two procedures are jointly learned to sort all nodes in the original graph, and then the top $k$ nodes are selected to form the new graph.

\subsubsection{Local Voting} \label{sec:ITAS}
Max pooling operations on grid-like data are performed
in local regions. Inspired by this, we propose the
local voting to measure node importance within a local region, which
explicitly incorporates graph topology and local information.
Specifically, we compute the
similarity scores between each node and its neighboring nodes.
The score for a node $i$ is the mean value of the
similarity scores with its neighboring nodes. The resulting
score for a node indicates the similarity between this
node and its neighboring nodes. If a node has a high local voting
score, it can highly represent a local graph that consists of it
and its neighboring nodes.
Formally, given a graph $G=(A,H)$ with the adjacency matrix $A\in\mathbb{R}^{n\times n}$ and
the feature matrix $H\in\mathbb{R}^{n\times d}$,
the local voting is expressed as
\begin{equation}\label{eq:ITAS}
\begin{aligned}
&R = HH^T\in\mathbb{R}^{n\times n},\\
&\hat{R} = R\circ (\hat{D}^{-1}\hat{A})\in\mathbb{R}^{n\times n},\\
&\mathbf{y}^L= \mbox{softmax}(\frac{1}{n}\hat{R}\mathbf{1}_n)\in\mathbb{R}^n,
\end{aligned}
\end{equation}
where $\hat{A} = A+I$ is used to add self-loops,
$\hat{D}$ is the diagonal degree matrix with $\hat{D}_{ii} = \sum_j\hat{A}_{ij}$,
$\circ$ denotes the element-wise matrix
multiplication operation,
$\mathbf{1}_n\in\mathbb{R}^n$ denotes a $n$-dimensional vector with all ones,
and $\mbox{softmax}(\cdot)$ denotes an element-aware softmax operation.

Suppose $H$ is a matrix with node features, denoted as $H$=$\{\mathbf{h}_1,\mathbf{h}_2,\cdot \cdot \cdot,\mathbf{h}_n\}^T$. For node $i$, its feature is represented as  $\mathbf{h}_i\in\mathbb{R}^d$. We first compute the similarity matrix $R$ based
on node features. Specifically, for any node $i$ and node $j$, their similarity score $r_{ij}$ is obtained by the dot product between $\mathbf{h}_i$ and $\mathbf{h}_j$.
Since $R$ contains similarity scores for all node pairs in the graph,
we perform an element-wise matrix multiplication between the similarity matrix $R$
and the normalized adjacency matrix $\hat{D}^{-1}\hat{A}$, resulting a matrix $\hat{R}$
where similarity scores for non-connected nodes are all zeros.
To this end, each row vector $\hat{\mathbf{r}}_i$ in $\hat{R}$ represents
similarity scores between
the node $i$ and its neighborhood.
Next, for the node $i$, we average its all similarity scores to indicate its importance within its local neighbors. Finally, we apply softmax function to obtain the final score vector, denoted as $\mathbf{y}^L=[y^L_1, y^L_2, \cdot \cdot \cdot, y^L_n]^T$ and $\sum_{i}y^L_i =1$.

Our proposed local voting essentially encodes local information from
neighborhood and explicitly incorporates graph topology information.
The node with a higher score indicates it has higher similarities with its
neighbors and hence tends to be more important.
Typically, by scoring each node based on local information,
local voting achieves a
similar objective as the max pooling operation on images. Max pooling on grid-like data keeps the most salient features within a local region.
Similarly, our
local voting regulates local information and encourages to select the most important nodes within a local region
when forming the new graph.
Notably, our proposed local voting considers both graph topology information and node features. Node features are used to compute the closeness between different nodes. In addition, we have several ways to compute closeness between two vectors, such as inner product, concatenation, and Gaussian function. In this work, we choose to employ the inner product since it has been
proven as a simple yet effective solution~\cite{wang2018non}. In practice, we can employ learnable weights to make the model more powerful. Specifically, we can add a learnable weight matrix $W_r$ to the computation of interaction matrix $R$ that $R = HW_rH^T$.
Figure~\ref{fig:gap} provides an illustration of our proposed local voting operation.

\subsubsection{Global Voting} \label{sec:FTAS}
For each node, local voting essentially shows its ``local'' role within a local region.
However, different from grid-like data such as images, local neighborhood regions
within graphs are ``overlapped''. We can not rely on local salient features only
to perform graph pooling.
In this section, we introduce our proposed global voting
to evaluate ``global'' role for each node.
Intuitively, local voting shows the importance of a node within its neighborhood,
while global voting shows how important its neighbor region (including itself) is in the entire graph.
Typically, the neighbor region of a node can be treated as a sub-graph and the center of this sub-graph is the node itself. Such a sub-graph may also carry important information.
For example, in a graph representing a protein, atoms are
nodes and bonds are edges. Then a sub-graph contains an atom and
its neighboring atoms. Certain sub-graphs can represent vital units and indicate
the functionality of the entire protein. Hence, such information should be captured in graph pooling. We propose global voting to measure the importance of different subgraphs in the entire graph.
Formally,
the global voting is express as
\begin{equation}\label{eq:FTAS}
\begin{aligned}
&\hat{H} = \hat{D}^{-1}\hat{A}H\in\mathbb{R}^{n\times n},
&\mathbf{y}^G = \mbox{softmax}(\hat{H}\mathbf{p})\in\mathbb{R}^n,
\end{aligned}
\end{equation}
where $\mathbf{p}\in\mathbb{R}^d$ is a trainable projection vector,
and $\mbox{softmax}(\cdot)$ denotes an element-aware softmax operation.
We first aggregate neighbor information for each node,
through which the new features of a node represent the information of its sub-graph.
After that, we perform the scalar projection $\mathbf{p}$ on the structure-aggregated feature matrix $\hat{H}$
and obtain the score vector $\mathbf{y}^G$.
The projection vector $\mathbf{p}$ is learnable and shared by all nodes in the
graph. In this way, the element $y^G_i$ in $\mathbf{y}^G$ indicates the importance of node $i$ and its sub-graph based on
the globally shared vector.

Note that the computation of global voting is similar to the recently proposed SortPool~\cite{zhang2018end} and top-k pooling~\cite{gao2019graph}. However, we explicitly consider the topology information to generate $\hat{H}$ while SortPool and top-k pooling may ignore such information. Without topology information, the capacity of networks is restricted and the functionalities of sub-graphs may be neglected. Our proposed
global voting measures the node importance in a global manner and considers both node
features and graph topology information.

\subsubsection{Graph Pooling Layer} \label{sec:ourpooling}
The local voting measures the local similarities between each node
and its neighborhood, and the global voting
captures the importance of different
neighbor regions in the entire graph.
Altogether, we combine these two to measure the importance of different nodes so that topology information is explicitly considered. Specifically, the final sorting vector
$\mathbf{y}$ is computed as the summation of
$\mathbf{y}^L$ and $\mathbf{y}^G$. Then based on the sorting vector $\mathbf{y}$, we rearrange the order of all the nodes in the original graph and
sample the top $k$ important nodes to form a new graph.
Formally, our proposed TAP layer can be expressed as
\begin{equation}\label{eq:pool}
\begin{aligned}
&\mathbf{y}=\mathbf{y}^L+\mathbf{y}^G\in\mathbb{R}^n,
&\mbox{idx} = \mbox{rank}(\mathbf{y},k)\in\mathbb{R}^k,\\
&H^\prime = H_{\mbox{idx},\ :}\ \in\mathbb{R}^{k\times d},
&A^\prime = A_{\mbox{idx},\ \mbox{idx}}\ \in\mathbb{R}^{k\times k},
\end{aligned}
\end{equation}
where $k$ is a pre-defined number indicating the number of nodes in the output graph,
$\mbox{rank}(\mathbf{y},k)$
is the operation for node sampling, which
returns the indexes of top-k values in $\mathbf{y}$.
The idx is a list of indexes indicating the selected nodes. $H_{\mbox{idx},\ :}$ is a row-aware extractor on feature matrix
to obtain the feature matrix $H^\prime$ for the new graph. $A_{\mbox{idx},\ \mbox{idx}}$ produces the new adjacency matrix based on
the node connections in the original graph.
With the ranking procedure,
only the top-k important nodes are selected according to the values in $\mathbf{y}$.
Then the new
feature matrix $H^\prime$ and adjacency matrix $A^\prime$
are obtained following the extraction process.
Notably, our TAP layer introduces negligibly additional parameters, since
the only used parameters include
the linear transformation matrix $W_r$ and the projection vector $\mathbf{p}$.

\begin{figure*}[t]
    \centering
    \includegraphics[width=\textwidth]{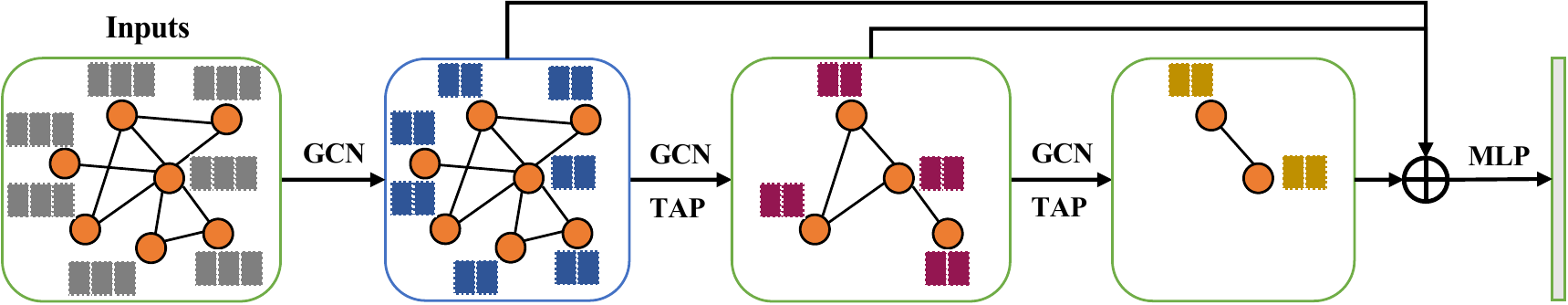}
    \caption{An illustration of the topology-aware pooling network. $\oplus$
    denotes the concatenation operation of feature vectors. Each node in the
    input graph contains three features. We use a GCN layer to transform the
    feature vectors into low-dimensional representations. We stack two blocks,
    each of which consists of a GCN layer and a TAP layer. A global reduction
    operation such as max-pooling is applied to the outputs of the first GCN
    layer and TAP layers. The resulting feature vectors are concatenated and fed
    into the final multi-layer perceptron for prediction.}\label{fig:model}
\end{figure*}

\subsection{Graph Connectivity Term}\label{sec:conterm}

Our proposed TAP layer computes the ranking scores by using
similarity scores between nodes in the graph, thereby regarding
topology information in the graph. However, the coarsened graph
generated by the TAP layer may suffer from the problem of
isolated nodes. In sparsely connected graphs, some nodes have a
very small number of neighboring nodes or even only themselves.
Suppose node $i$ only connects to itself. The local voting score of
node $i$ is the similarity score to itself, which may result in
high local voting scores, thereby generating high ranking scores in the graph. The resulting graph can be very
sparsely connected, which completely loses the original graph
structure. In the extreme situation, the coarsened graph can
contain only isolated nodes without any connectivity. This can
inevitably hurt the model performance.

To overcome the limitation of TAP layer and encourage better
connectivity in the selected graph, we propose to add a graph
connectivity term to the computation of ranking scores. To this
end, we use node degrees as an indicator for graph connectivity
and add degree values to their ranking scores such that
densely-connected nodes are preferred during nodes selection. By
using the node degree as the graph connectivity term, the ranking
score of node $i$ is computed as
\begin{equation}
s_i = y_i + \lambda \frac{d_i}{n},
\label{eq:conterm}
\end{equation}
where $y_i$ is the ranking score achieved in Eq.~(\ref{eq:pool}) of node $i$, $d_i$ is the degree of node $i$, and $\lambda$ is a
hyper-parameter that sets the importance of the graph
connectivity term to the computation of ranking scores. The graph
connectivity term can overcome the limitation of the TAP layer.
The computation of ranking scores now considers nodes degrees and
gives rise to better connectivity in the resulting graph. A
better connected coarsened graph is expected to retain more graph
structure information, thereby leading to better model
performance.

\subsection{Topology-Aware Pooling Networks}\label{sec:tapn}

Based on our proposed TAP layer, we build a family of networks
known as topology-aware pooling networks~(TAPNets) for graph
classification tasks. In TAPNets, we firstly apply a graph
embedding layer to produce low-dimensional representations of
nodes in the graph, which helps to deal with some datasets with
very high-dimensional input feature vectors. Here, we use a GCN
layer~\cite{kipf2016semi} for node embedding. After the embedding
layer, we stack several blocks, each of which consists of a GCN
layer for high-level feature extraction and a TAP layer for graph
coarsening. In the $i^{th}$ TAP layer, we use a hyper-parameter
$k^{(i)}$ to control the number of nodes in the sampled graph. We
feed the output feature matrices of the graph embedding layer and
TAP layers to a classifier.

In TAPNets, we use a multi-layer perceptron as the classifier. We
first transform network outputs to a one-dimensional vector.
Global
max and average pooling operations are two popular ways for the
transformation, which reduce the spatial size of feature matrices
to 1. Recently, \cite{xu2018powerful} proposed to use the
summation function that results in promising performances. In
TAPNets, we concatenate transformation output vectors produced by
the global pooling operations using max, averaging, and
summation, respectively. The resulting feature vector is fed into
the classifier. Figure~\ref{fig:model} illustrates a sample
TAPNet with two blocks.


\subsection{Auxiliary Link Prediction Objective}\label{sec:aux}

Multi-task learning has shown to be effective across various
machine learning
tasks~\cite{ruder2017overview,ying2018hierarchical}. It can
leverage useful information in multiple related tasks, thereby
leading to better generalization and performance. In this
section, we propose to add an auxiliary link prediction objective
during training by using a by-product of our TAP layer. In
Eq.~(\ref{eq:ITAS}), we compute the similarity scores
$R$ between every pair of nodes in the graph. By
applying an element-wise $\mbox{softmax}(\cdot)$ on
$R$, we can obtain a link probability matrix
$\tilde{R}\in \mathbb{R}^{n \times n}$ with
each element $\tilde{r}_{ij}$ measures the likelihood of a link
between node $i$ and node $j$ in the graph. With the adjacency
matrix $A$, we compute the auxiliary link
prediction loss as
\begin{equation}
    loss_{aux} = \frac{1}{n^2} \sum_{i=1}^{n} \sum_{j=1}^{n}
    f\left(\tilde{R}_{ij},A_{ij}\right),
\end{equation}
where $f(\cdot, \cdot)$ is a loss function that computes the
distance between the link probability matrix
$\tilde{R}$ and the adjacency matrix
$A$.

Note that the adjacency matrix used as the link prediction
objective is directly derived from the original graph. Since the
TAP layer extracts a sub-graph from the original one, the
connectivity between two nodes in the sampled graph is the same
as that in the original graph. This means the adjacency matrices
in deeper networks are still using the original graph structure.
Compared to auxiliary link prediction in
DiffPool~\cite{ying2018hierarchical} that uses the learned
adjacency matrix as objective, our method uses the original
links, thereby providing more accurate information. This can also
be clearly observed in the experimental studies in
Sections~\ref{sec:cmp_soc} and~\ref{sec:cmp_bio}.

\begin{table*}[t] \caption{Comparisons between TAPNets and other models
    in terms
    of graph classification accuracy~(\%) on social network datasets
    including COLLAB, IMDB-BINARY, IMDB-MULTI, REDDIT-BINARY, REDDIT-MULTI5K
    and REDDIT-MULTI12K datasets.}\label{table:large_cmp}
\small
\begin{tabularx}{\textwidth}{l Y Y Y Y Y c}\hline
    \textbf{}              & \textbf{COLLAB}         & \textbf{IMDB-B}         & \textbf{IMDB-M}         & \textbf{RDT-B}          & \textbf{RDT-M5K}        & \textbf{RDT-M12K}       \\ \hline\hline
    \# graphs              & 5000                    & 1000                    & 1500                    & 2000                    & 4999                    & 11929                   \\ \hline
    \# nodes               & 74.5                    & 19.8                    & 13.0                    & 429.6                   & 508.5                   & 391.4                   \\ \hline
    \# classes             & 3                       & 2                       & 3                       & 2                       & 5                       & 11                      \\ \hline
    \hline
    WL~\cite{shervashidze2011weisfeiler}      & 78.9 $\pm$ 1.9          & 73.8 $\pm$ 3.9          & 50.9 $\pm$ 3.8          & 81.0 $\pm$ 3.1          & 52.5 $\pm$ 2.1          & 44.4 $\pm$ 2.1          \\ \hline
    PSCN~\cite{niepert2016learning}           & 72.6 $\pm$ 2.2          & 71.0 $\pm$ 2.2          & 45.2 $\pm$ 2.8          & 86.3 $\pm$ 1.6          & 49.1 $\pm$ 0.7          & 41.3 $\pm$ 0.8          \\ \hline
    DGCNN~\cite{zhang2018end}                 & 73.8                    & 70.0                    & 47.8                    & -                       & -                       & 41.8                    \\ \hline
    DIFFPOOL~\cite{ying2018hierarchical}      & 75.5                    & -                       & -                       & -                       & -                       & 47.1                    \\ \hline
    g-U-Net~\cite{gao2019graph}               & 77.5 $\pm$ 2.1          & 75.4 $\pm$ 3.0          & 51.8 $\pm$ 3.7          & 85.5 $\pm$ 1.3          & 48.2 $\pm$ 0.8          & 44.5 $\pm$ 0.6          \\ \hline
    GIN~\cite{xu2018powerful}                 & 80.6 $\pm$ 1.9          & 75.1 $\pm$ 5.1          & 52.3 $\pm$ 2.8          & 92.4 $\pm$ 2.5          & \textbf{57.5 $\pm$ 1.5} & -                       \\ \hline
    \textbf{TAPNet}~(ours)                    & \textbf{84.9 $\pm$ 1.8} & \textbf{79.9 $\pm$ 4.5} & \textbf{56.2 $\pm$ 3.1} & \textbf{94.6 $\pm$ 1.8} & 57.3 $\pm$ 1.5          & \textbf{49.4 $\pm$ 1.7} \\ \hline
    \hline
\end{tabularx}
\end{table*}

\section{Experimental Studies}

In this section, we evaluate our methods and networks on graph
classification tasks using bioinformatics and social network
datasets. We conduct ablation experiments to evaluate the
contributions of the TAP layer and each term in it to the overall
network performance.

\subsection{Experimental Setup}
We evaluate our methods using social network datasets and bioinformatics
datasets. They share the same experimental setups except for minor differences.
The node features in social network networks are created using one-hot encodings
of node degrees. The nodes in bioinformatics have categorical features. We use
the TAPNet proposed in Section~\ref{sec:tapn} that consists of one GCN layer and
three blocks. The first GCN layer is used to learn low-dimensional
representations of nodes in the graph. Each block is composed of one GCN layer
and one TAP layer. All GCN and TAP layers output 48 feature maps. We use Leaky
ReLU~\cite{xu2015empirical} with a slop of 0.01 to activate the outputs of GCN
layers. The three TAP layers in the networks select numbers of nodes that are
proportional to the nodes in the graph. We use the rates of 0.8, 0.6, and 0.4 in
three TAP layers, respectively. We use $\lambda = 0.1$ to control the importance
of the graph connectivity term in the computation of ranking scores.
Dropout~\cite{srivASTava2014dropout} is applied to the input feature matrices
of GCN and TAP layers with keep rate of 0.7. We use a two-layer feed-forward
network as the network classifier. Dropout with keep rate of 0.8 is applied to
input features of two layers. We use ReLU activation function on the output of
the first layer on DD, PTC, MUTAG, COLLAB, REDDIT-MULTI5K, and REDDIT-MULTI12K
datasets. We use ELU~\cite{clevert2015fAST} for other datasets. We train our
networks using Adam optimizer~\cite{kingma2014adam} with a learning rate of
0.001. To avoid over-fitting, we use $L_2$ regularization with $\lambda=0.0008$.
All models are trained using one NVIDIA GeForce RTX 2080 Ti GPU.

We compare our method with several state-of-the-art baselines.
Weisfeiler-Lehman subtree kernel (WL)~\cite{shervashidze2011weisfeiler}
is treated as the most effective kernel method for graph representation learning.
PSCN~\cite{niepert2016learning} learns node representations from neighborhood and uses a canonical node ordering
for graph representations.
DGCNN~\cite{zhang2018end} applies several GCN layers and proposes SortPool to perform graph pooling by sorting and selecting nodes.
DiffPool~\cite{ying2018hierarchical} is developed based on GraphSage~\cite{hamilton2017inductive} and proposes a hierarchical pooling technique, which learns to perform node clustering to build the new graph.
g-U-Net~\cite{gao2019graph} proposes top-$k$ pooling that employs a projection vector to compute the rank score for each node. The graph topology is not considered when computing rank scores. SAGPool~\cite{Lee2019self} is similar to top-$k$ pooling and encodes topology information. 
The graph connectivity issue is not tackled in SAGPool.
GIN~\cite{xu2018powerful} is the graph isomorphism network,
whose representational power is shown to be similar to the power of the WL test.
EigenPool~\cite{ma2019graph} is another node-clustering method
that is based on graph Fourier
transform and utilizes local structures
when performing node clustering.
HaarPool~\cite{wang2020haarpooling} employs Haar basis and the compressive Haar transforms
to generate a sparse characterization of the
input graph while preserving structure information.

\begin{table*}[t] \caption{Comparisons between TAPNets and other models in terms of
    graph classification accuracy~(\%) on bioinformatics datasets including
    DD, PTC, MUTAG, and PROTEINS datasets.}\label{table:small_cmp} \small
\begin{tabularx}{\textwidth}{l Y Y Y Y}\hline
    \textbf{}              & \textbf{DD}             & \textbf{PTC}            & \textbf{MUTAG}          & \textbf{PROTEINS}       \\ \hline\hline
    \# graphs              & 1178                    & 344                     & 188                     & 1113                    \\ \hline
    \# nodes               & 284.3                   & 25.5                    & 17.9                    & 39.1                    \\ \hline
    \# classes             & 2                       & 2                       & 2                       & 2                       \\ \hline
    \hline
    WL~\cite{shervashidze2011weisfeiler}                     & 78.3 $\pm$ 0.6          & 59.9 $\pm$ 4.3          & 90.4 $\pm$ 5.7          & 75.0 $\pm$ 3.1          \\ \hline
    PSCN~\cite{niepert2016learning}                   & 76.3 $\pm$ 2.6          & 60.0 $\pm$ 4.8          & 92.6 $\pm$ 4.2          & 75.9 $\pm$ 2.8          \\ \hline
    DGCNN~\cite{zhang2018end}                  & 79.4 $\pm$ 0.9          & 58.6 $\pm$ 2.4          & 85.8 $\pm$ 1.7          & 75.5 $\pm$ 0.9          \\ \hline
    SAGPool~\cite{Lee2019self}                & 76.5                    & -                       & -                       & 71.9                    \\ \hline
    DIFFPOOL~\cite{ying2018hierarchical}               & 80.6                    & -                       & -                       & 76.3                    \\ \hline
    g-U-Net~\cite{gao2019graph}                & 82.4 $\pm$ 2.9          & 64.7 $\pm$ 6.8          & 87.2 $\pm$ 7.8          & 77.6 $\pm$ 2.6          \\ \hline
    GIN~\cite{xu2018powerful}                    & 82.0 $\pm$ 2.7          & 64.6 $\pm$ 7.0          & 90.0 $\pm$ 8.8          & 76.2 $\pm$ 2.8          \\ \hline
    EigenPool~\cite{ma2019graph}              & 0.786          & -          & 0.801          & 0.766 \\ \hline
    HaarPool~\cite{wang2020haarpooling}       & -          & -          & 90.0 $\pm$ 3.6         & 80.4 $\pm$ 1.8 \\ \hline    
    \textbf{TAPNet}~(ours) & \textbf{84.6 $\pm$ 3.5} & \textbf{73.4 $\pm$ 5.8} & \textbf{93.8 $\pm$ 6.0} & \textbf{80.8 $\pm$ 4.5} \\ \hline
    \hline
\end{tabularx}
\end{table*}

\subsection{Graph Classification Results on Social Network Datasets}\label{sec:cmp_soc}
We conduct experiments on graph classification tasks to evaluate
our proposed methods and TAPNets. We use 6 social network
datasets; those are COLLAB, IMDB-BINARY, IMDB-MULTI,
REDDIT-BINARY, REDDIT-MULTI5K and
REDDIT-MULTI12K~\cite{yanardag2015structural} datasets. Note
that REDDIT datasets are popular large datasets used for network
embedding learning in terms of graph size and number of
graphs~\cite{ying2018hierarchical,xu2018powerful}. Since there
is no feature for nodes in social networks, we create node
features by following the practices in~\cite{xu2018powerful}. In
particular, we use one-hot encodings of node degrees as feature
vectors for nodes in social network datasets. On these datasets,
we perform 10-fold cross-validation as in~\cite{zhang2018end}
with 9 folds for training and 1 fold for testing. To ensure fair
comparisons, we do not use the auxiliary link prediction
objective in these experiments. We compare our
TAPNets with other state-of-the-art models in terms of graph
classification accuracy. The comparison results are summarized in
Table~\ref{table:large_cmp}. We can observe from the results that
our TAPNets significantly outperform previous best models on most social
network datasets by margins of 4.3\%, 4.8\%, 3.9\%, 2.2\%, and
2.3\% on COLLAB, IMDB-BINARY, IMDB-MULTI, REDDIT-BINARY, and
REDDIT-MULTI12K datasets, respectively. The promising
performances, especially on large datasets such as REDDIT,
demonstrate the effectiveness of our methods. Note that the
superior performances over g-U-Net~\cite{gao2019graph} show that
our TAP layer can produce better-coarsened graph than that using
the top-$k$ pooling layer.

\subsection{Graph Classification Results on Bioinformatics Datasets}\label{sec:cmp_bio}

To fully evaluate our methods, we conduct experiments on graph
classification tasks using 4 bioinformatics datasets; those are
DD
~\cite{dobson2003distinguishing}
, PTC
~\cite{toivonen2003statistical}
, MUTAG
~\cite{wale2008comparison}
, and PROTEINS
~\cite{borgwardt2005protein}
~\cite{xu2018powerful} datasets.
Notably, nodes in bioinformatics datasets have categorical
features. In these experiments, we do not use the auxiliary link
prediction objective. We compare our TAPNets with other
state-of-the-art models in terms of graph classification accuracy
without using the auxiliary link prediction term in loss
function. The comparison results are summarized in
Table~\ref{table:small_cmp}. We can observe from the results that
our TAPNets achieve significantly better results than other
models by margins of 2.2\%, 8.7\%, 3.8\%, and 0.4\% on DD, PTC,
MUTAG, and PROTEINS datasets, respectively. Notably, some
bioinformatics datasets such as PTC and MUTAG are much smaller
than social network datasets in terms of number of graphs and
number of nodes in graphs. The promising results on these small
datasets demonstrate that our methods can achieve good
generalization and performances without the risk of over-fitting.
Also, the superior performances over SAGPool on DD and PROTEINS
datasets demonstrate that our methods can better capture the
topology information.


\subsection{Comparison with Other Graph Pooling Layers}\label{sec:cmp_pools}

\begin{table}
    \caption{Comparisons between different pooling operations
        based on the same TAPNet architecture
        in terms of the graph classification accuracy~(\%)
        on PTC, IMDB-MULTI, and REDDIT-BINARY datasets.
        }\label{table:pool_cmp}\small
    \begin{tabularx}{\linewidth}{l Y c Y}\hline
        \textbf{Model}       & \textbf{PTC}  & \textbf{IMDB-M} & \textbf{RDT-B} \\ \hline\hline
        Net$_{\mbox{diff}}$  & 70.9          & 54.9            & 92.1           \\ \hline
        Net$_{\mbox{sort}}$  & 70.6          & 54.8            & 92.3           \\ \hline
        Net$_{\mbox{top-}k}$ & 71.5          & 55.2            & 92.8           \\ \hline
        \textbf{TAPNet}      & \textbf{73.4} & \textbf{56.2}   & \textbf{94.6}  \\ \hline
        \hline
    \end{tabularx}
    \vspace{-8pt}
\end{table}

It may be argued that our TAPNets achieve promising results by employing
superior networks. In this section, we conduct experiments on the same TAPNet
architecture to compare our TAP layer with other graph pooling layers; those are
DIFFPOOL, SortPool, and top-$k$ pooling layers. We denote the networks with
the TAPNet architecture while using these pooling layers as Net$_{\mbox{diff}}$,
Net$_{\mbox{sort}}$, and Net$_{\mbox{top-}k}$, respectively. We evaluate them on
PTC, IMDB-MULTI, and REDDIT-BINARY datasets and summarize the results in
Table~\ref{table:pool_cmp}. Note that these models use the same experimental
setups to ensure fair comparisons. The results demonstrate the superior
performance of our proposed TAP layer compared with other pooling layers using
the same network architecture.

\subsection{Ablation Studies}

\begin{table}
    \caption{Comparisons among TAPNets with and without TAP
        layers, TAPNet without local voting term~(LV), TAPNet without global voting term~(GV), TAPNet without graph connection term~(GCT),
        and TAPNet with auxiliary link prediction objective~(AUX)
        in terms of the graph classification accuracy~(\%) on
        PTC, IMDB-MULTI, and REDDIT-BINARY
        datasets.}\label{table:aba_cmp}
    \small
    \begin{tabularx}{\linewidth}{l Y c c}\hline
        \textbf{Model}        & \textbf{PTC}  & \textbf{IMDB-M} & \textbf{RDT-B} \\ \hline\hline
        TAPNet~w/o~TAP        & 70.6          & 52.1            & 91.0           \\ \hline
        TAPNet~w/o~LV        & 72.1          & 55.3           & 93.2           \\ \hline
        TAPNet~w/o~GV        & 72.7          & 55.6            & 94.1           \\ \hline
        TAPNet~w/o~GCT        & 73.1          & 55.8            & 94.2          \\ \hline
        TAPNet                & 73.4         & 56.2            & 94.6          \\ \hline
        \textbf{TAPNet w AUX} & \textbf{73.7} & \textbf{56.4}   & \textbf{94.8}  \\ \hline
        \hline
    \end{tabularx}
    \vspace{-8pt}
\end{table}

\begin{figure*}[t]
    \centering
    \includegraphics[width=0.8\textwidth]{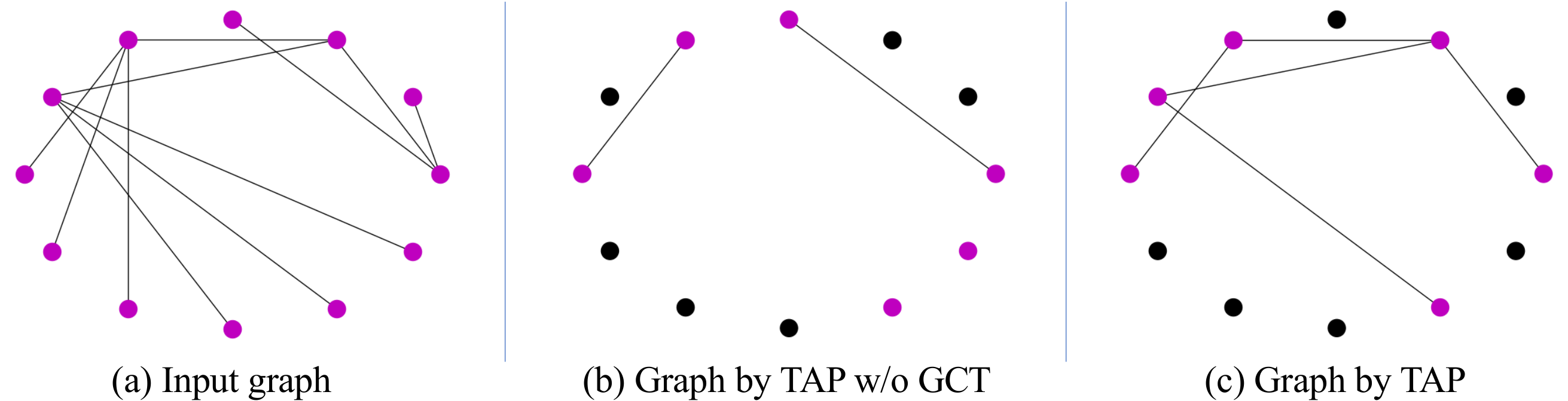}
    \caption{Visualization of coarsened graphs by TAP and
        TAP~w/o~GCT. Here, GCT denotes the graph connection term.
        Based on the input graph~(a), the pooling layers select 6
        nodes to form new graphs. The nodes that are not selected
        are colored black. The new graph in~(b) generated by
        TAP~w/o~GCT is sparsely connected. The one generated by
        TAP in~(c) is shown to be better connected.}
    \label{fig:gs}
\end{figure*}

In this section, we investigate the contributions of TAP layer
and its components in ranking score computation; those are the
local voting term~(LV), the global voting term~(GV)and the graph connectivity term~(GCT).
We remove TAP layers from TAPNet which we denote as
TAPNet~w/o~TAP. To explore the contributions of terms in ranking
scores computation, we separately remove LVs, GVs and GCTs from all
TAP layers in TAPNets. We denote the resulting models as
TAPNet~w/o~LV, TAPNet~w/o~GV and TAPNet~w/o~GCT, respectively. In addition, we
add the auxiliary link prediction objective as described in
Section~\ref{sec:aux} in training. We denote the TAPNet using
auxiliary training objective as TAPNet~w~AUX. We evaluate these
models on three datasets; those are PTC, IMDB-MULTI, and
REDDIT-BINARY datasets.

The comparison results on these datasets are summarized in
Table~\ref{table:aba_cmp}. The results show that TAPNets
outperform TAPNets w/o TAP by margins of 2.8\%, 4.1\%, and 3.5\%
on PTC, IMDB-MULTI, and REDDIT-BINARY datasets, respectively. The
better results of TAPNet over TAPNet~w/o~SST and TAPNet~w/o~GCT
show the contributions of LVs, GVs, and GCTs to performances. It can
be observed that TAPNet~w~AUX achieves better performances than
TAPNet, which shows the effectiveness of the auxiliary link
prediction objective. To fully study the impact of GCT on TAP
layer, we visualize the coarsened graphs generated by TAP and TAP
without GCT~(denoted as TAP~w/o~GCT). We select a graph from PTC
dataset and illustrate outcome graphs in Figure~\ref{fig:gs}. We
can observe from the figure that TAP produces a
better-connected graph than that by TAP~w/o~GCT.

\subsection{Parameter Study of TAP}

\begin{table}
    \caption{Comparisons among TAPNets with and without TAP layers, and TAPNet
        without local voting and global voting~(LV\&GV)
        in terms of the graph classification
        accuracy~(\%), and the number of parameters
        on REDDIT-BINARY dataset.}
    \label{table:param_cmp}\small
    \begin{tabularx}{\linewidth}{@{\hskip3pt}l @{\hskip4pt} Y @{\hskip4pt} Y @{\hskip4pt} Y@{\hskip3pt}}\hline
        \textbf{Model} & \textbf{Accuracy} & \textbf{\#Params} & \textbf{Ratio} \\ \hline\hline
        TAPNet w/o TAP & 91.0              & 323,666           & 0.00\%         \\ \hline
        TAPNet w/o LV\&GV & 91.5              & 323,666           & 0.00\%         \\ \hline
        TAPNet         & \textbf{94.6}     & 331,090           & 2.29\%         \\ \hline
        \hline
    \end{tabularx}
    \vspace{-6pt}
\end{table}

Since TAP layer employs linear transformation to compute
ranking scores, it involves extra trainable parameters to the
overall network. Here, we conduct experiments to study the number
of parameters in TAPNet. We remove the extra trainable parameters
from TAP layers in two ways; those are removing TAP layers from
the TAPNet and removing similarity score terms LV and GV from TAP
layers. We denote the resulting two networks as TAPNet~w/o~TAP
and TAPNet~w/o~LV\&GV, respectively.
The comparison results on REDDIT-BINARY dataset is summarized in
Table~\ref{table:param_cmp}. We can see from the results that TAP
layers only need 2.29\% additional trainable parameters. We
believe the negligible usage of extra parameters will not
increase the risk of over-fitting but can bring 3.6\% and 3.1\% performance
improvement over TAPNet~w/o~TAP and TAPNet~w/o~LV\&GV on
REDDIT-BINARY dataset. Also, the promising performances of
TAPNets on small datasets like PTC and MUTAG in
Table~\ref{table:small_cmp} show that TAP layers will not
significantly increase the number of trainable parameters or
cause the over-fitting problem.

\subsection{Performance Study of \texorpdfstring{$\lambda$}{Lg}}

\begin{figure}
    \includegraphics[width=\linewidth]{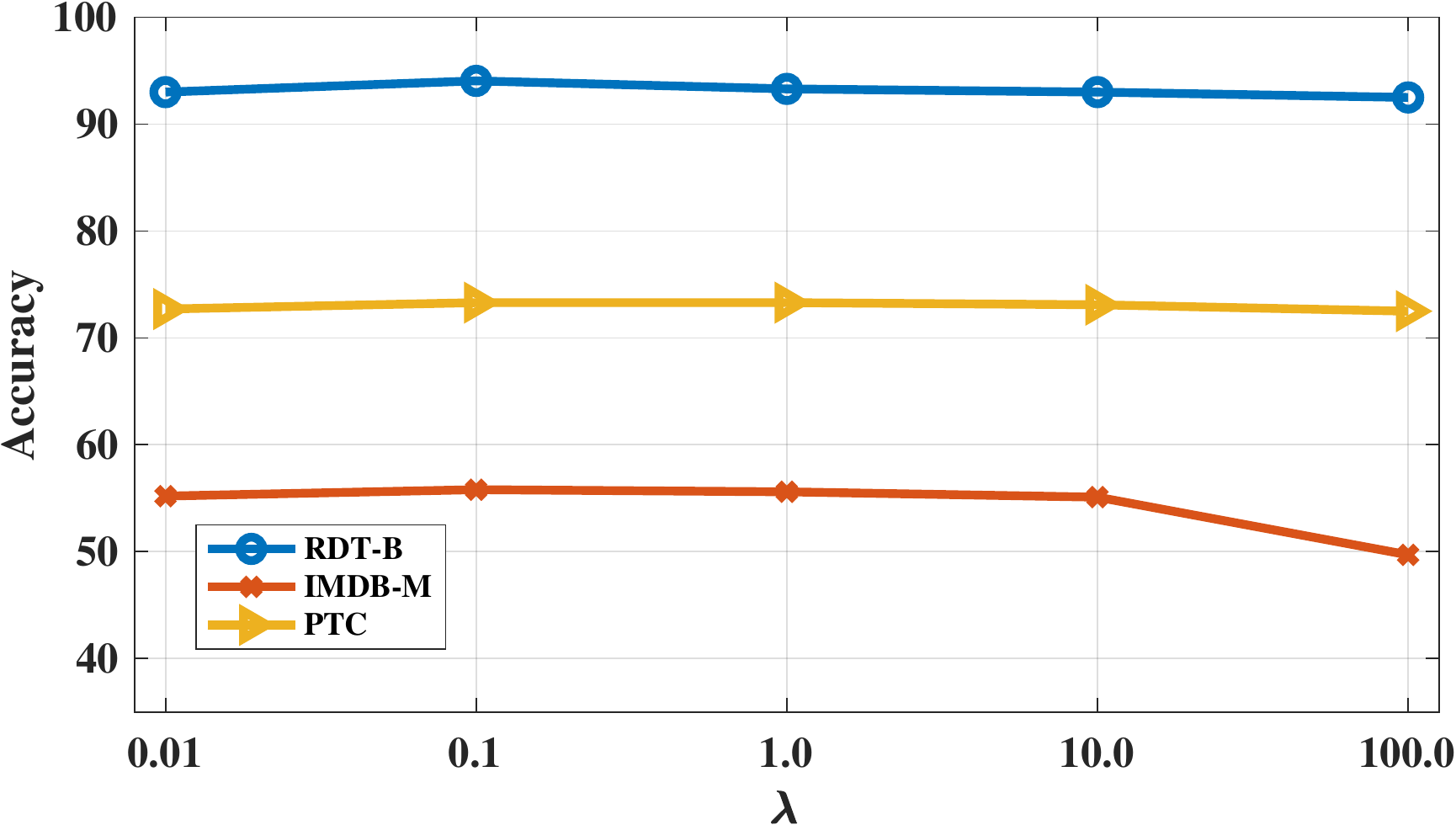}
    \caption{Comparison results of TAPNets using different
    $\lambda$ values in TAP layers. We report graph
    classification accuracies on PTC, IMDB-BINARY, and
    REDDIT-BINARY datasets.}
    \vspace{-8pt}
    \label{fig:lambda}
\end{figure}

In Section~\ref{sec:conterm}, we propose to add the graph
connectivity term into the computation of ranking scores to
improve the graph connectivity in the coarsened graph. It can be
seen that $\lambda$ is an influential hyper-parameter in the TAP
layer. In this part, we study the impacts of different $\lambda$
values on network performances. We select different $\lambda$
values from the range of 0.01, 0.1, 1.0, 10.0, and 100.0 to cover
a reasonable range of values. We evaluate TAPNets using different
$\lambda$ values on PTC, IMDB-MULTI, and REDDIT-BINARY datasets.

The results are shown in Figure~\ref{fig:lambda}. We can observe
that the best performances on three datasets are achieved with
$\lambda=0.1$. When $\lambda$ becomes larger, the performances of
TAPNet models decrease. This indicates that the graph
connectivity term is a plus term for generating reasonable ranking
scores but it should not overwhelm the similarity score term that
encodes the topology information in ranking scores.

\section{Conclusions}

In this work, we propose a novel topology-aware pooling~(TAP)
layer that explicitly encodes the topology information in ranking
scores. A TAP layer performs a two-stage voting process that
includes the local voting and
global voting.
The local voting
attends each node to its neighboring nodes
and uses the average similarity score with its neighboring nodes
as its local importance score. 
The global voting employs a projection vector to compute 
a similarity score for each node as its global importance score.
Then the final ranking score for a node is the combination of 
its local and global voting scores.
The node sampling based on these
ranking scores explicitly incorporates the topology information, thereby
leading to a better-coarsened graph. Moreover, we propose to add
a graph connectivity term to the computation of ranking scores to
overcome the isolated problem which a TAP layer may suffer from.
Based on the TAP layer, we develop topology-aware pooling
networks~(TAPNets) for network representation learning. We add an
auxiliary link prediction objective to train our networks by
employing the similarity score matrix generated in TAP layers.
Experimental results on graph classification tasks using both
bioinformatics and social network datasets demonstrate that our
TAPNets achieve performance improvements as compared to previous
models. Ablation Studies show the contributions of our TAP layers
to network performance.

\ifCLASSOPTIONcompsoc
  \section*{Acknowledgments}
\else
  \section*{Acknowledgment}
\fi

This work was supported by National Science Foundation grant
IIS-2006861.

\end{document}